\documentclass[a4paper]{article}

\usepackage{INTERSPEECH2020}

\usepackage{lipsum}
\usepackage{bm,amsmath,amssymb,mathrsfs,amsfonts}
\usepackage{caption}
\usepackage{scalefnt} 
\usepackage{multirow} 
\usepackage{enumitem}
\usepackage{type1cm}
\usepackage{cite}
\usepackage{url}

\AtBeginDocument{
  \abovedisplayskip     =0.5\abovedisplayskip
  \abovedisplayshortskip=0.5\abovedisplayshortskip
  \belowdisplayskip     =0.5\belowdisplayskip
  \belowdisplayshortskip=0.5\belowdisplayshortskip}

\title{CTC-synchronous Training for Monotonic Attention Model}

\name{Hirofumi Inaguma$^1$, Masato Mimura$^1$, Tatsuya Kawahara$^1$}
\address{
  $^1$Graduate School of Informatics, Kyoto University, Kyoto, Japan}
\email{\{inaguma,mimura,kawahara\}@sap.ist.i.kyoto-u.ac.jp}

\begin{document}

\maketitle
%
\begin{abstract}
Monotonic chunkwise attention (MoChA) has been studied for the online streaming automatic speech recognition (ASR) based on a sequence-to-sequence framework.
In contrast to connectionist temporal classification (CTC), backward probabilities cannot be leveraged in the alignment marginalization process during training due to left-to-right dependency in the decoder.
This results in the error propagation of alignments to subsequent token generation.
To address this problem, we propose CTC-synchronous training (CTC-ST), in which MoChA uses CTC alignments to learn optimal monotonic alignments.
Reference CTC alignments are extracted from a CTC branch sharing the same encoder with the decoder. 
The entire model is jointly optimized so that the expected boundaries from MoChA are synchronized with the alignments.
Experimental evaluations of the TEDLIUM release-2 and Librispeech corpora show that the proposed method significantly improves recognition, especially for long utterances. 
We also show that CTC-ST can bring out the full potential of SpecAugment for MoChA.
\end{abstract}

\section{Introduction}\label{sec:intro}
\vspace{-1mm}
Streaming automatic speech recognition (ASR) is a core technology used in simultaneous interpretation such as live captioning, simultaneous translation, and dialogue systems.
Recently, the performance of end-to-end (E2E) ASR systems has been nearing that of hybrid systems with much more simplified architectures \cite{google_sota_asr,karita2019comparative}.
Therefore, building effective streaming E2E-ASR systems is an important step towards making E2E models applicable in the real world.

The streaming E2E models proposed thus far can be categorized as \textit{time-synchronous} or \textit{label-synchronous} models. 
Time-synchronous models include connectionist temporal classification (CTC) \cite{ctc_graves}, recurrent neural network transducer (RNN-T) \cite{rnn_transducer}, and recurrent neural aligner (RNA) \cite{recurrent_neural_aligner}.
These models can make predictions from the input stream frame-by-frame, so they can easily satisfy the demands of streaming ASR.
Label-synchronous models based on an attention-based sequence-to-sequence (S2S) framework \cite{chorowski2015attention} outperform time-synchronous models on several benchmarks in offline scenarios \cite{s2s_comparison_google,s2s_comparison_baidu,huang2019exploring}.
However, label-synchronous models are not suitable for streaming ASR because the initial token cannot be generated until all of the speech frames in an utterance are encoded.
To make label-synchronous models streamable, several variants have been studied, such as hard monotonic attention \cite{hard_monotonic_attention}, monotonic chunkwise attention (MoChA) \cite{mocha}, triggered attention \cite{moritz2019triggered_icassp2019}, adaptive computation steps \cite{adaptive_computation_steps}, and continuous integrate-and-fire \cite{cif}.
In this work, we focus on MoChA as a streaming attention-based S2S model since it can be trained efficiently and shows promising results for various ASR tasks \cite{kim2020attention,online_hybrid_ctc_attention,adaptive_mocha,inaguma2020streaming,online_hybrid_ctc_attention_taslp2020}.

\begin{figure}[t]
  \centering
  \includegraphics[width=0.95\linewidth]{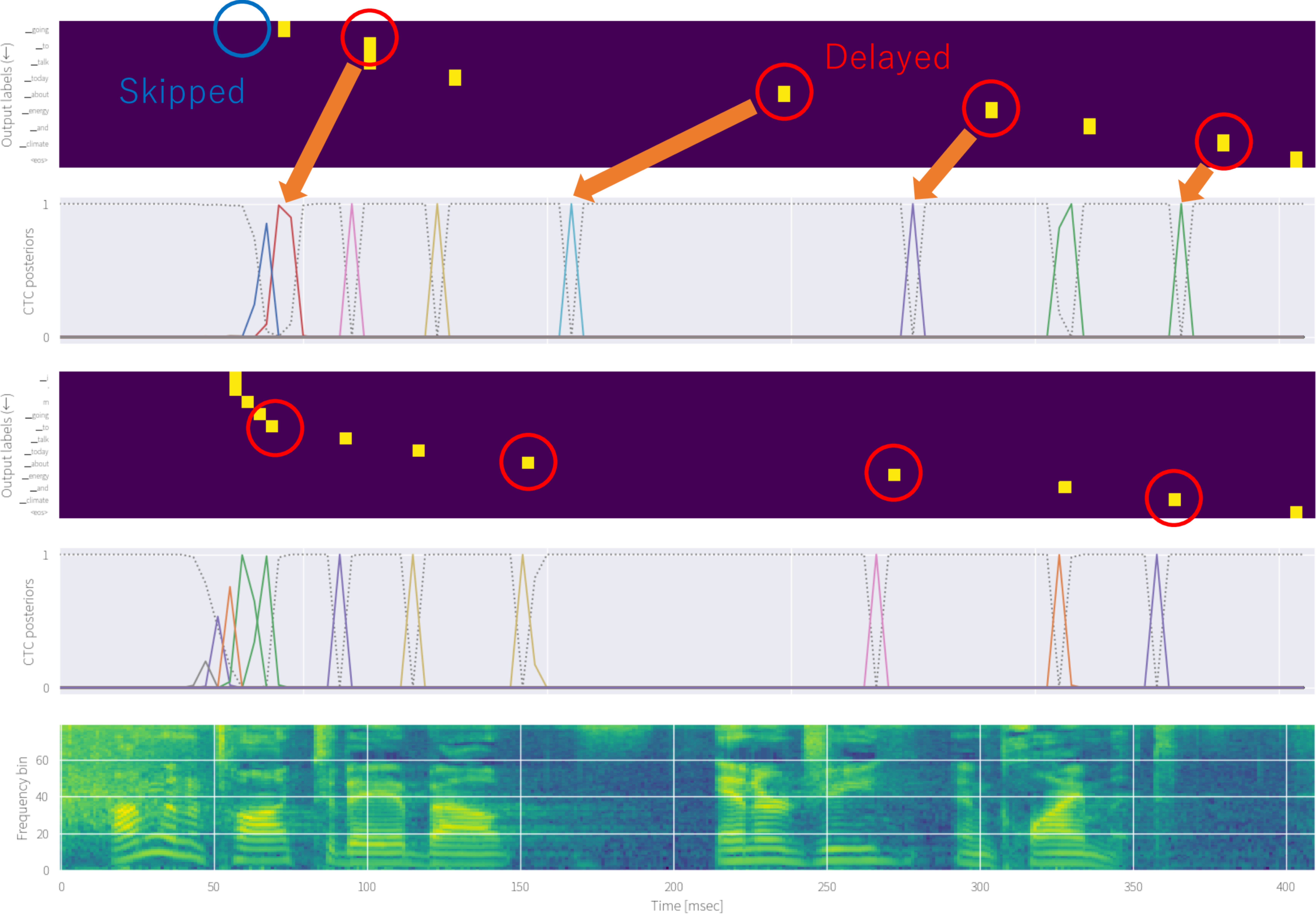}
  \vspace{-2.5mm}
  \caption{Visualization of decision boundaries (yellow dots) of LC-BLSTM-40+40 - MoChA (top, T5) and proposed model w/ CTC-ST (bottom, T6)}
  \label{fig:attention_plot}
  \vspace{-7mm}
\end{figure}

Hard monotonic attention \cite{hard_monotonic_attention} enables online linear-time decoding during inference by introducing a discrete binary variable.
To train the model using standard backpropagation, the alignment probabilities are marginalized over all possible paths during training.
MoChA is an extension of hard monotonic attention and introduces additional soft attention over a small chunk. 
This lessens the strict monotonic constraints of input-output alignments and leads to more accurate recognition \cite{mocha}.
However, the results of training MoChA have often been reported to be unstable \cite{online_hybrid_ctc_attention,adaptive_mocha,inaguma2020streaming,online_hybrid_ctc_attention_taslp2020}.
This is because the attention scores are not globally normalized over all encoder memories, which results in poorly scaled attention scores in the early training stage. 
Consequently, the gap in attention behaviors widens between training and test time.
A possible solution to this problem is to regularize the model so that the total attention scores over all output timesteps sum up to the output sequence length.
This was originally proposed in \cite{cif,inaguma2020streaming} and referred to as \textit{quantity loss}. 
Although \cite{inaguma2020streaming} reported that quantity loss is not effective for MoChA for large-scale data (3.4k hours), we have found that it is very effective when the training data size is small ($<$ 1k hours).
However, apart from the scaling issue, alignments in MoChA are susceptible to noise from previous timesteps because they are calculated only with the forward algorithm due to left-to-right dependency in the autoregressive decoder.
This is because the backward algorithm cannot be used to estimate accurate alignments, unlike in CTC.
This leads to significant error propagation of the predicted token boundary positions to subsequent token generation, especially for long utterances.

In this work, we propose \textit{CTC-synchronous training} (CTC-ST) to learn reliable monotonic alignments by using CTC alignments.
Since CTC is optimized with the forward-backward algorithm, the resulting alignments are more reliable than those from MoChA.
Moreover, the CTC posterior probabilities tend to peak in spikes \cite{ctc_graves}, so they can be regarded as decision boundaries for token generation and a good reference for MoChA to learn the appropriate timing to generate output tokens.
We extract the reference boundaries from the CTC model with forced alignment, and the expected decision boundaries from MoChA are optimized to be close to the corresponding CTC boundaries.
Since we jointly optimize the MoChA and CTC branches by sharing the same encoder, the entire model can be trained in an end-to-end fashion.
We also propose a curriculum training strategy to stabilize the training.

We experimentally evaluated the TEDLIUM release-2 and Librispeech corpora to demonstrate that the proposed CTC-ST significantly improves recognition.
We also investigate reliability of CTC alignments in the proposed CTC-ST by combining it with SpecAugment \cite{specaugment}.

\vspace{-2mm}
\section{Related work}\label{sec:related_work}
\vspace{-1mm}
CTC alignments are used in triggered attention for streaming E2E-ASR \cite{moritz2019triggered_icassp2019,moritz2019streaming_asru2019,moritz2020streaming}, in which encoder memories are truncated with every CTC spike and the global attention mechanism is applied subsequently.
Our work differs in that (1) we do not use CTC spikes in the test phase, and (2) the decoding complexity of MoChA is linear while that of the triggered attention is quadratic but streamable.

Regarding manipulating boundary positions in hard monotonic attention, \cite{inaguma2020streaming} uses the external framewise alignments extracted from the hybrid ASR system to reduce latency for token generation in MoChA.
In contrast, our proposed CTC-ST does not require any external framewise alignments and is designed to improve recognition.

\vspace{-2mm}
\section{Streaming attention-based S2S}
\vspace{-1mm}
\subsection{Latency-controlled BLSTM (LC-BLSTM)}\label{ssec:latency_controlled}
\vspace{-1mm}
In order to enable low-latency feature encoding while maintaining bidirectionality, latency-controlled bidirectional long short-term memory (LC-BLSTM) encoder has been introduced to restrict future frames to small chunks ($N_{\rm r}$ frames) \cite{latency_controlled_blstm,xue2017improving,online_hybrid_ctc_attention,adaptive_mocha}.
LC-BLSTM consists of forward and backward LSTMs similar to the BLSTM but processes a small chunk by sliding a window of size $N_{\rm c}$ without overlap between adjacent chunks.
The backward LSTM processes later $N_{\rm r}$ frames as future contexts in addition to $N_{\rm c}$ frames in the current chunk.
Thus, the total latency at each chunk is $(N_{\rm c} + N_{\rm r})$ frames.
The backward LSTM state is reset at every chunk while the previous state of the forward LSTM is carried over to the next chunk as the initial state.
We refer to this as LC-BLSTM-$N_{\rm c}$+$N_{\rm r}$ in the later experiments.

\vspace{-2mm}
\subsection{Monotonic chunkwise attention (MoChA)}\label{ssec:mocha}
\vspace{-1mm}
To enable online linear-time decoding for attention-based S2S models during inference, hard monotonic attention was proposed by introducing discrete binary decision processes \cite{hard_monotonic_attention}.
Unlike the global attention mechanism \cite{chorowski2015attention}, all attention scores are assigned to a single memory at each output timestep $i$ without global score normalization.
As hard attention is not differentiable, the expected alignment probabilities are marginalized over all possible paths during training by introducing a selection probability $p_{i,j}\in[0,1]$.
$p_{i,j}$ is a function of the monotonic energy activation $e_{i,j}$, which is parameterized with the $i$-th decoder state $s_{i}$ and the $j$-th encoder state $h_{j}$ as follows:
%
\begin{eqnarray}
\alpha_{i,j} &=& p_{i,j}\sum_{k=1}^{j}\bigg(\alpha_{i-1,k}\prod_{l=k}^{j-1}(1-p_{i,l})\bigg) \nonumber \\
&=& p_{i,j}\bigg((1-p_{i,j-1})\frac{\alpha_{i,j-1}}{p_{i,j-1}}+\alpha_{i-1,j}\bigg) \label{eq:mocha_alpha} \\
e_{i,j} &=& g\frac{v^{\mathsf T}}{||v||}{\rm ReLU}({\mathbf W}_{\rm h}h_{j} + {\mathbf W}_{\rm s}s_{i} + b) + r \label{eq:mocha_e_mono} \\
p_{i,j} &=& \mbox{Sigmoid}(e_{i,j}) \nonumber
\end{eqnarray}
where $g$, $v$, ${\mathbf W}_{\rm h}$, ${\mathbf W}_{\rm s}$, $b$, and $r$ are learnable parameters.

Monotonic chunkwise attention (MoChA) \cite{mocha} is an extension of the above method which introduces additional soft chunkwise attention to loosen the strict input-output alignment in hard attention.
The chunkwise attention scores $\beta_{i,j}$ over the small $w$ frames are calculated from each expected boundary $j$:
%
\begin{gather}
\beta_{i,j} = \sum_{k=j}^{j+w-1}\bigg(\alpha_{i,k}{\rm exp}(u_{i,j})/\sum_{l=k-w+1}^{k}{{\rm exp}(u_{i,l})}\bigg) \label{eq:mocha_beta}
\end{gather}
where $u_{i,j}$ is the chunk energy activation formulated similar to $e_{i,j}$ in Eq. \eqref{eq:mocha_e_mono} without weight normalization and the offset $r$.
The $i$-th context vector is calculated as a weighted sum of encoder memories by $\beta_{i,j}$ and the subsequent token generation processes are the same as in the global attention.
Both $\alpha_{i,j}$ and $\beta_{i,j}$ can be calculated in parallel with the cumulative sum/product and the moving sum operations.

At the time of the test, each token is generated once $p_{i,j}$ exceeds a threshold of 0.5.
The next token boundary is determined further to the right than the current boundary.
For more details on the decoding algorithm, refer to \cite{hard_monotonic_attention,mocha}.

\vspace{-2mm}
\subsection{Quantity regularization}\label{ssec:quantity_loss}
\vspace{-1mm}
Since MoChA does not normalize monotonic attention scores $\alpha_{i,j}$ across all encoder memories $\{h_{j}\}_{j=1}^{T}$, there is no guarantee that $\sum_{j=1}^{T}{\alpha_{i,j}} = 1$ is satisfied during training.
This results in poorly scaled selection probabilities $p_{i,j}$ and widens the gap in behaviors between training and testing because $\alpha_{i,j}$ can attenuate quickly during marginalization.
To address this drawback, we use a simple regularization term $\mathcal{L}_{\rm qua}$, i.e., \textit{quantity loss} \cite{cif,inaguma2020streaming}, to encourage the total number of the expected boundaries to be close to the reference sequence length $U$: $\mathcal{L}_{\rm qua} = |U - \sum_{i=1}^{U}{\sum_{j=1}^{T}{\alpha_{i,j}}}|$.
Scaling $\alpha_{i,j}$ properly is expected to encourage the discreteness of $p_{i,j}$ during training, which also leads to better estimation of $\beta_{i,j}$ in Eq. \eqref{eq:mocha_beta}.
The objective function is designed with the interpolation of the negative log-likelihood $\mathcal{L}_{\rm mocha}$, CTC loss $\mathcal{L}_{\rm ctc}$, and quantity loss $\mathcal{L}_{\rm qua}$ with tunable interpolation weights $\lambda_{\rm ctc}$ ($0 \le \lambda_{\rm ctc} \le 1$) and $\lambda_{\rm qua}$ ($\ge 0$) as follows:
%
\begin{gather}
\mathcal{L}_{\rm total} = (1- \lambda_{\rm ctc}) \mathcal{L}_{\rm mocha} + \lambda_{\rm ctc} \mathcal{L}_{\rm ctc} + \lambda_{\rm qua} \mathcal{L}_{\rm qua} \label{eq:total_loss_v1}
\end{gather}
%
We perform joint optimization with the CTC objective by sharing the encoder sub-network to encourage monotonicity of the input-output alignment \cite{hybrid_ctc_attention}.


\vspace{-2mm}
\section{CTC-synchronous training (CTC-ST)}\label{sec:ctc_sync}
\vspace{-1mm}
Hard monotonic attention in MoChA depends entirely on past alignments because of the left-to-right dependency in Eq. \eqref{eq:mocha_alpha}.
Although the scaling of the monotonic attention scores $\alpha_{i,j}$ can be facilitated with quantity regularization described in Section \ref{ssec:quantity_loss}, the alignment errors in the middle and latter steps cannot be recovered, which leads to the significant error propagation of $\alpha_{i,j}$ to the latter tokens as the output sequence becomes longer.
Unlike CTC, the decoder is autoregressive, so it is difficult to apply the backward algorithm during the alignment marginalization process.
Consequently, the decision boundaries tend to shift to the right side (future) from the corresponding reference acoustic boundaries \cite{adaptive_mocha,kim2020attention,inaguma2020streaming}.

On the other hand, since CTC marginalizes all possible alignments with the forward-backward algorithm during training, its decision boundaries are more reliable and tend to shift to the left compared to those of MoChA.
We indeed observed a delay of a few frames between the decision boundaries from the initial MoChA model and the corresponding CTC spikes when sharing the same encoder (see Figure \ref{fig:attention_plot}).
This also arises a problem that hard monotonic attention activates at the different position over the shared encoder outputs from the corresponding CTC spike.
Meanwhile, the well-trained CTC posterior probability distributions tend to peak in sharp spikes \cite{ctc_graves}.
Therefore, the decision boundaries from CTC are expected to serve as an effective guide for MoChA to learn accurate alignments.
In other words, MoChA can correct error propagation from past decision boundaries with the help of the CTC alignments.

In this work, we propose \textit{CTC-synchronous training} (CTC-ST) to provide reliable CTC alignments as references to MoChA for learning robust monotonic alignments.
The MoChA model is trained to mimic the CTC model to generate the similar decision boundaries.
Both the MoChA and CTC branches are jointly optimized by sharing the same encoder, and the decision boundaries as the reference are extracted from the CTC branch.
Thus, we can train the model in an end-to-end fashion without the use of external alignments, unlike in \cite{inaguma2020streaming}.
Synchronizing both decision boundaries can be regarded as explicit interaction between MoChA and CTC on the decoder side.

We use the most probable CTC path $\bar{\pi}$ of length $T$ after forced alignment with the forward-backward algorithm \cite{moritz2019triggered_icassp2019} and use time indices of non-blank labels in $\bar{\pi}$ as the reference boundary positions ${\mathbf b}^{\rm ctc}=({\rm b}^{\rm ctc}_{1},\ldots,{\rm b}^{\rm ctc}_{U})$.
Note that the leftmost index is used when non-blank labels are repeated without interleaving any blank labels.
For the end-of-sentence token, the last input index (i.e., $T$) is used.
${\mathbf b}^{\rm ctc}$ is generated with model parameters at each training step on-the-fly and updates as the training continues.
The objective function of the CTC-ST $\mathcal{L}_{\rm sync}$ is defined as follows:
%
%
\begin{eqnarray*}\label{eq:latency_loss}
\mathcal{L}_{\rm sync} = \frac{1}{U} \sum_{i=1}^{U}|{\rm b}^{\rm ctc}_{i} - {\rm b}^{\rm mocha}_{i}|
\end{eqnarray*}
where ${\rm b}^{\rm mocha}_{i}=\sum_{j=1}^{T}j\alpha_{i,j}$ is the expected decision boundary of MoChA for the $i$-th token during training.
The total objective function in Eq. \eqref{eq:total_loss_v1} is modified accordingly as follows:
\begin{multline}
\mathcal{L}_{\rm total} = (1- \lambda_{\rm ctc}) \mathcal{L}_{\rm mocha} + \lambda_{\rm ctc} \mathcal{L}_{\rm ctc} \\ 
+ \lambda_{\rm qua} \mathcal{L}_{\rm qua} + \lambda_{\rm sync} \mathcal{L}_{\rm sync} \label{eq:total_loss_v2}
\end{multline}
where $\lambda_{\rm sync}$ ($\ge 0$) is a tunable parameter, set to 1.0 in this work.
Unless otherwise noted, $\lambda_{\rm qua}$ is set to 0 when using CTC-ST.


\vspace{-2.5mm}
\subsection{Curriculum learning strategy}\label{ssec:curriculum_learning}
\vspace{-1mm}
Since CTC-ST is designed to bring decision boundaries from both MoChA and CTC closer, their alignment probabilities must be peaky to minimize $\mathcal{L}_{\rm sync}$ in Eq. \eqref{eq:latency_loss}.
In the early training stage, however, monotonic attention scores $\alpha_{i,j}$ tend to be diffused over several frames and are not normalized to sum up to one.
Applying CTC-ST from scratch is ineffective, so we use a curriculum learning strategy instead.
First, the MoChA model with a standard BLSTM encoder is trained with random initialization together with Eq. \eqref{eq:total_loss_v1} until convergence ({\bf stage-1}).
Next, after loading model parameters in stage-1, future contexts are restricted for the LC-BLSTM encoder and the parameters are optimized with CTC-ST by Eq. \eqref{eq:total_loss_v2} ({\bf stage-2}).
Although we mainly use the LC-BLSTM encoder, this two-staged training can also be applied to the MoChA model with the unidirectional LSTM encoder by using the same encoder in both stages.

\vspace{-2.5mm}
\subsection{Combination with SpecAugment}\label{ssec:specaugment}
\vspace{-1mm}
SpecAugment has been shown to greatly enhance the decoder in the global attention model by performing on-the-fly data augmentation \cite{specaugment}.
However, since it introduces time and frequency masks into the input log-mel spectrogram, the recurrency in Eq. \eqref{eq:mocha_alpha} can be easily collapsed after the masked region.
In our experiments, MoChA was not shown to improve as the global attention model with SpecAugment.
In contrast, CTC-ST can recover $\alpha_{i,j}$ after the masked region with CTC spikes since CTC is formulated on the conditional independence assumption per frame.
Therefore, CTC-ST is beneficial for MoChA to learn the monotonic alignments that withstand noisy inputs.
We will also analyze the impact of mask size.

\begin{table}[t]
    \centering
    \footnotesize
    \begingroup
    \caption{TEDLIUM2 results. ${}^{\clubsuit}$Quantity regularization is used.}\label{tab:result_tedlium2}
    \vspace{-3mm}
    \begin{tabular}{c|l|c} \toprule
     \multicolumn{2}{c|}{Model} & $\%$WER \\ \hline
        \multirow{7}{*}{\rotatebox{90}{Offline}} 
          & LSTM - Global attention & 11.9 \\
          & BLSTM - Global attention ({\tt T1}) & 9.5 \\
          & \ + LC-BLSTM-40+20 (seed: {\tt T1}) & 10.1 \\
          & \ + LC-BLSTM-40+40 (seed: {\tt T1}) & 9.7 \\
          & BLSTM - MoChA & 12.6 \\
          & \ + Quantity regularization ({\tt T2}) & {\bf 9.8} \\
          & \ + CTC-ST & 10.2 \\ \hline \hline
        \multirow{6}{*}{\rotatebox{90}{Streaming}} 
          & LSTM - MoChA${}^{\clubsuit}$ ({\tt T3}) & 15.0 \\
          & \ + CTC-ST ({\tt T4}, seed: {\tt T3}) & {\bf 13.2} \\ \cline{2-3}
          & LC-BLSTM-40+20 - MoChA${}^{\clubsuit}$ (seed: {\tt T2}) & 12.2 \\
          & \ + CTC-ST & {\bf 10.5} \\ \cline{2-3}
          & LC-BLSTM-40+40 - MoChA${}^{\clubsuit}$ ({\tt T5}, seed: {\tt T2}) & 11.3 \\
          & \ + CTC-ST ({\tt T6}) & {\bf 9.9} \\ \bottomrule
    \end{tabular}
    \vspace{-5mm}
    \endgroup
\end{table}

\vspace{-2mm}
\section{Experiments}\label{sec:exp}
\vspace{-1mm}
\subsection{Experimental setup}
\vspace{-1mm}
We used the TEDLIUM release-2 (210 hours, lecture) \cite{tedlium} and Librispeech (960 hours, reading) \cite{librispeech} corpora for experimental evaluations.
We extracted 80-channel log-mel filterbank coefficients computed with a 25-ms window size shifted every 10 ms using Kaldi \cite{kaldi}.
We performed 3-fold speed perturbation \cite{speed_perturbation} on the TEDLIUM2 corpus with factors of 0.9, 1.0, and 1.1.

The encoders were composed of two CNN blocks followed by five layers of (LC-)BLSTM.
Each CNN block was composed of two layers of CNN having a $3 \times 3$ filter followed by a max-pooling layer with a stride of $2 \times 2$, which resulted in $4$-fold frame rate reduction.
We set the number of cells in each (LC-)BLSTM layer to $512$ per direction.
We summed up the LSTM outputs in both directions at each layer to reduce the input dimension of the subsequent (LC-)BLSTM layer \cite{tuske2019advancing}.
The memory cells were doubled when using the unidirectional LSTM encoder.
The decoder was a single layer of unidirectional LSTM with $1024$-dimensional memory cells.
For offline models, we used the location-based attention \cite{chorowski2015attention}.
We set the chunk size $w$ of MoChA to $4$, which was tuned in our preliminary experiments.
$r$ in Eq. \eqref{eq:mocha_e_mono} was initialized with $-4$.
We used 10k vocabularies based on the Byte Pair Encoding (BPE) algorithm \cite{sennrich2015neural}.

Optimization was performed using Adam \cite{adam} with learning rate $1e-3$ and it was exponentially decayed.
We used dropout and label smoothing \cite{label_smoothing} with probabilities $0.4$ and $0.1$, respectively.
$\lambda_{\rm ctc}$ was set to $0.3$.
We set $\lambda_{\rm qua}$ to $2.0$ and $0.01$ on the TEDLIUM2 and Librispeech corpora, respectively.
We used a $4$-layer LSTM language model (LM) with $1024$ memory cells for decoding with a beam width of $10$ \cite{shallow_fusion}.
Scores were normalized by the number of tokens for MoChA.
CTC scores were not leveraged during inference.\footnote{We implemented models with Pytorch \cite{pytorch}. Detailed hyperparameter settings during training and decoding are available at \url{https://github.com/hirofumi0810/neural_sp}.}

\vspace{-2mm}
\subsection{Results}\label{ssec:result}
\vspace{-1mm}
Table \ref{tab:result_tedlium2} shows the results for the TEDLIUM2.
For offline models, MoChA ({\tt T2}) approached the performance of the global attention model ({\tt T1}).
Quantity regularization was essential for attaining suitable performance for the baseline MoChA.
CTC-ST also improved the performance by a large margin, but it was less effective than quantity regularization when applying from scratch.
This is because the scale of $\alpha_{i,j}$ was not adequate in the early training stage.
For streaming models, our proposed CTC-ST with the curriculum learning strategy significantly improved the performances of the unidirectional LSTM, LC-BLSTM-40+20, and LC-BLSTM-40+40 MoChA models by $12.0$, $13.9$, and $12.3$\%, respectively.
Note that CTC-ST was not combined with quantity regularization in stage-2.
Although a larger future context was helpful for boosting performance, the effectiveness of CTC-ST was orthogonal.



\begin{table}[t]
    \centering
    \footnotesize
    \begingroup
    \caption{Results of curriculum learning on the TEDLIUM2}\label{tab:result_tedlium2_curriculum}
    \vspace{-3mm}
    \begin{tabular}{c|l|cc|cc} \toprule
      \multicolumn{2}{c|}{\multirow{2}{*}{Model}} & \multirow{2}{*}{\shortstack{Quantity\\regularization}} & \multirow{2}{*}{CTC-ST} & \multirow{2}{*}{$\%$WER} \\
      \multicolumn{2}{c|}{} & & &  \\ \hline
    \multirow{5}{*}{\rotatebox{90}{Streaming}} 
      & \multirow{1}{*}{LC-BLSTM-40+40} & \checkmark & -- & 12.3 \\ \cline{2-5}
      & \multirow{4}{*}{\shortstack{\ + Curriculum\\learning\\(seed: {\tt T2})}} & -- & -- & 16.9 \\
      &  & \checkmark & -- & 11.3 \\
      & & -- & \checkmark & {\bf 9.9} \\
      & & \checkmark & \checkmark & 10.1 \\ \bottomrule
    \end{tabular}
    \vspace{-2.5mm}
    \endgroup
\end{table}

\begin{table}[t]
    \centering
    \footnotesize
    \begingroup
    \caption{Results with SpecAugment on the TEDLIUM2}\label{tab:result_tedlium2_specaugment}
    \vspace{-3mm}
    \begin{tabular}{c|l|cc|c} \toprule
     \multicolumn{2}{c}{Model} & \multicolumn{1}{|c}{$F$} & $T$ & $\%$WER \\ \hline
     \multirow{4}{*}{\rotatebox{90}{Offline}} 
       & Transformer \cite{karita2019comparative} & 30 & 40 & 8.1 \\
       & BLSTM - Global attention \cite{zeyer2019comparison} & N/A & N/A & 8.8 \\ \cline{2-5}
       & \multirow{2}{*}{BLSTM - Global attention} & - & - & 9.5 \\ 
       &  & 27 & 100 & 8.1 \\ \hline \hline
    \multirow{7}{*}{\rotatebox{90}{Streaming}} 
       & \multirow{3}{*}{\shortstack{LC-BLSTM-40+40 - MoChA\\(seed: {\tt T2})}} & - & - & 11.3 \\
       & & 27 & 100 & 12.8 \\
       & & 13 & 50 & 11.2 \\ \cline{2-5}
       & \multirow{4}{*}{\ + CTC-ST} & - & - & 9.9 \\
       & & 27 & 100 & 9.0 \\
       & & 27 & 50 & {\bf 8.6} \\
       & & 13 & 50 & 9.0 \\ \bottomrule
    \end{tabular}
    \vspace{-2.5mm}
    \endgroup
\end{table}

\begin{figure}[t]
  \centering
  \includegraphics[width=0.99\linewidth]{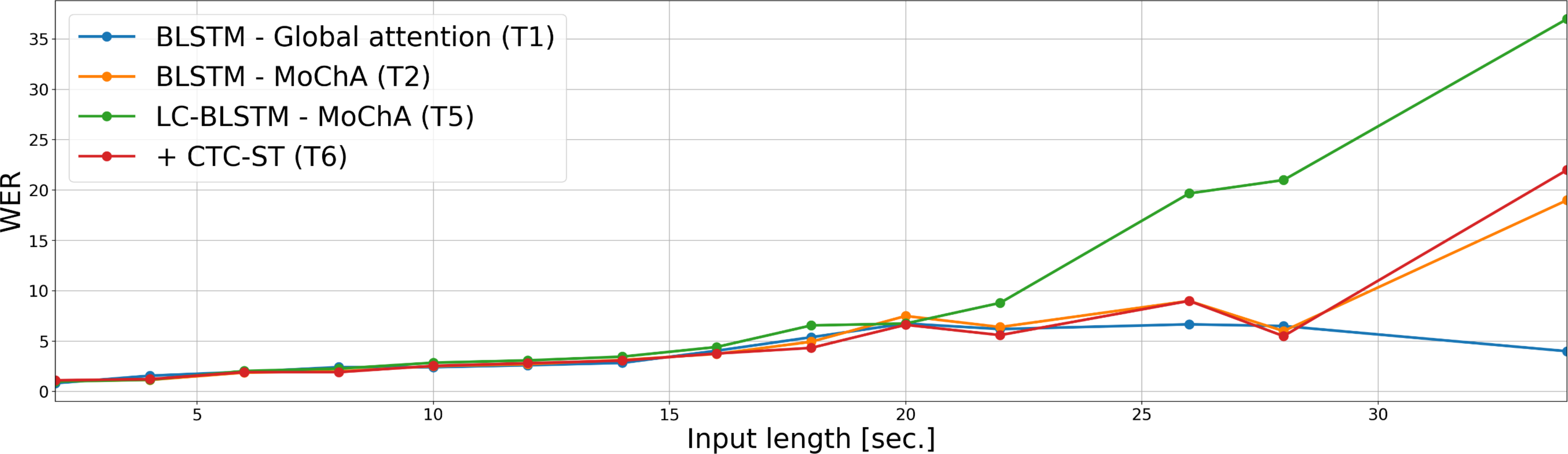}
  \vspace{-3mm}
  \caption{WER performance bucketed by different input length on the TEDLIUM2}
  \label{fig:wer_dist}
  \vspace{-5.5mm}
\end{figure}

\begin{table}[t]
    \centering
    \footnotesize
    \begingroup
    \caption{Librispeech results. ${}^{\clubsuit}$Quantity regularization is used. $^{\diamondsuit}$SpecAugment is used.}\label{tab:result_librispeech}
    \vspace{-3mm}
    \begin{tabular}{c|l|cc} \toprule
     \multicolumn{2}{c|}{\multirow{2}{*}{Model}} & \multicolumn{2}{|c}{$\%$WER} \\ \cline{3-4}
      \multicolumn{2}{c|}{} & clean & other \\ \hline
      \multirow{5}{*}{\rotatebox{90}{Offline}} 
        & BLSTM - Global attention$^{\diamondsuit}$ \cite{karita2019comparative} & 3.3 & 10.8 \\ \cline{2-4}
        & BLSTM - Global attention ({\tt L1}) & 3.1 & 9.5  \\
        & \ + SpecAugment (seed: {\tt L1}) & 2.8 & 7.6 \\
        & BLSTM - MoChA & 3.6 & 10.5 \\ 
        & \ + Quantity regularization ({\tt L2}) & {\bf 3.3} & {\bf 10.0} \\ \hline
        
      \multirow{16}{*}{\rotatebox{90}{Streaming}} 
        & Transformer - CIF$^{\diamondsuit}$ \cite{cif} & 3.3  & 9.7 \\
        & Transformer - Triggered attention$^{\diamondsuit}$ \cite{moritz2020streaming} & 2.8 & 7.2 \\
        & PTDLSTM - Triggered attention \cite{moritz2019streaming_asru2019} & 5.9 & 16.8 \\
        & LSTM - MoChA + MWER$^{\diamondsuit}$ \cite{kim2020attention} & 5.6 & 15.6 \\
        & LSTM - MoChA + \{char, BPE\}-CTC \cite{garg2019improved} & 4.4 & 15.2 \\
        & LC-BLSTM - sMoChA \cite{online_hybrid_ctc_attention} & 6.0 & 16.7 \\
        & LC-BLSTM - MTA \cite{online_hybrid_ctc_attention_taslp2020} & 4.2 & 12.3 \\
        \cline{2-4}
        
        & LSTM - MoChA${}^{\clubsuit}$ ({\tt L3}) & 5.3 & 14.5 \\
        & \ + CTC-ST (seed: {\tt L3}) & 4.7 & 13.6 \\
        & \ ++ SpecAugment ($F$=13, $T$=50) & {\bf 4.2} & {\bf 11.2} \\ \cline{2-4}
        & LC-BLSTM-40+40 - MoChA${}^{\clubsuit}$ (seed: {\tt L2}) & 4.1 & 11.2 \\
        & \ + SpecAugment${}^{\clubsuit}$ ($F$=27, $T$=100) & 5.0 & 9.7 \\
        & \ + SpecAugment${}^{\clubsuit}$ ($F$=13, $T$=50) & 4.0 & 9.5 \\ 
        & \ + CTC-ST & 3.9 & 11.2 \\ 
        & \ ++ SpecAugment ($F$=27, $T$=100) & 3.6 & 9.2 \\
        & \ ++ SpecAugment ($F$=27, $T$=50) & {\bf 3.5} & {\bf 9.1} \\ 
        & \ ++ SpecAugment ($F$=13, $T$=50) & 3.6 & 9.4 \\ \bottomrule
    \end{tabular}
    \vspace{-6mm}
    \endgroup
\end{table}

Next, we investigated the effectiveness of regularization terms and the curriculum learning strategy for the LC-BLSTM-MoChA, shown in Table \ref{tab:result_tedlium2_curriculum}.
We used BLSTM-MoChA ({\tt T2}) as a seed model except for the first row and optimized the model with either CTC-ST, quantity regularization, or both.
Curriculum learning was highly effective and CTC-ST ({\tt T6}) significantly outperformed the case using quantity regularization ({\tt T5}).
Combining the two did not lead to any further improvements although it was more effective than the model with quantity regularization only.
This is likely because CTC-ST has an effect to encourage the proper scale of $\alpha_{i,j}$ in hard monotonic attention similarly to quantity regularization.

The combination of CTC-ST and SpecAugment is shown in Table \ref{tab:result_tedlium2_specaugment}.
We used two time masks with time mask parameter $T$ and two frequency masks with frequency mask parameter $F$ in \cite{specaugment}.
We applied SpecAugment to MoChA in stage-2 only because applying SpecAugment from scratch did not converge.
SpecAugment did not improve the performance of the MoChA models without the guide from CTC alignments because the attention scores in Eq. \eqref{eq:mocha_alpha} can be easily collapsed as mentioned in Section \ref{ssec:specaugment}.
CTC-ST solved this issue and led to an additional $13.1$\% relative improvement.
Moreover, CTC-ST was robust to input mask size.

We plotted the WERs bucketed by input length in Figure \ref{fig:wer_dist}.
The largest CTC-ST gains came from long utterances.
The offline global attention model ({\tt T1}) did not have difficulty in recognizing long utterances, whereas the initial LC-BLSTM MoChA model ({\tt T5}) did.
The proposed CTC-ST mitigated this problem ({\tt T6}).
The decision boundaries from the MoChA and CTC branches extracted from {\tt T3} and {\tt T4} are visualized in Figure \ref{fig:attention_plot}.
We found that the gap between the two boundaries was reduced, and the CTC spikes slightly shifted to the left.
This is also beneficial for reducing user perceived latency \cite{inaguma2020streaming}.

Finally, the results for Librispeech are shown in Table \ref{tab:result_librispeech}.
When using the unidirectional LSTM encoder for MoChA, we obtained $11.3$\% and $6.2$\% relative improvements with CTC-ST on the test-clean and test-other sets, respectively.
SpecAugment further improved the performance.
As the training data size is much larger, the gains from quantity regularization for the BLSTM encoder and CTC-ST for the LC-BLSTM encoder were smaller than those in TEDLIUM2, although both were still beneficial.
CTC-ST was effective for obtaining gains from SpecAugment and led to additional improvements of $10.2$\% and $18.7$\% on the test-clean and test-other sets, respectively.
The model parameter size of LC-BLSTM-MoChA was $53.77$M and fixed through all experiments.
Our optimal model requires only $860$ms (= $400$ms ($N_{\rm c}$) + $400$ms ($N_{\rm r}$) + $60$ms (CNN)) lookahead frames and the decoding complexity is linear.


\vspace{-2mm}
\section{Conclusion}\label{sec:conclusion}
\vspace{-1mm}
We proposed CTC-synchronous training (CTC-ST) to provide hard monotonic attention in MoChA with reliable alignments extracted from CTC.
By jointly training both sub-networks with the shared encoder and generating CTC alignments simultaneously, we enabled effective interaction between MoChA and CTC.
Experimental evaluations revealed that CTC-ST significantly improved the performance of MoChA and greatly reduced the gap from the offline models.
Further gains with SpecAugment were obtained when CTC-ST was applied, thus verifying its robustness to noisy alignments.

\bibliographystyle{IEEEtran}
\bibliography{reference}

\end{document}